\documentclass[a4paper, 10pt, conference]{ieeeconf}
\usepackage{amsmath}
\usepackage{graphicx}
\usepackage{xcolor}
\graphicspath{{./figures/}}
\usepackage{subfiles}
\usepackage{multirow}
\usepackage{amssymb}
\usepackage{booktabs}
\usepackage{algorithm}
\usepackage{algpseudocode}
\IEEEoverridecommandlockouts

\title{\bf Evaluating Gaussian Grasp Maps for Generative Grasping Models}
\author{
\authorblockN{William Prew\authorrefmark{1}\authorrefmark{2}, Toby P. Breckon\authorrefmark{1}, Magnus Bordewich\authorrefmark{1}, Ulrik Beierholm\authorrefmark{2},
\thanks{\authorrefmark{1}Department of Computer Science, Durham University, UK.}
\thanks{\authorrefmark{2}Department of Psychology, Durham University, UK.}
\thanks{Email: william.t.prew@durham.ac.uk}
}}

\begin{document}

\maketitle

\begin{abstract}
Generalising robotic grasping to previously unseen objects is a key task in general robotic manipulation. The current method for training many antipodal generative grasping models rely on a binary ground truth grasp map generated from the centre thirds of correctly labelled grasp rectangles. However, these binary maps do not accurately reflect the positions in which a robotic arm can correctly grasp a given object. We propose a continuous Gaussian representation of annotated grasps to generate ground truth training data which achieves a higher success rate on a simulated robotic grasping benchmark. Three modern generative grasping networks are trained with either binary or Gaussian grasp maps, along with recent advancements from the robotic grasping literature, such as discretisation of grasp angles into bins and an attentional loss function. Despite negligible difference according to the standard rectangle metric, Gaussian maps better reproduce the training data and therefore improve success rates when tested on the same simulated robot arm by avoiding collisions with the object: achieving 87.94\% accuracy. Furthermore, the best performing model is shown to operate with a high success rate when transferred to a real robotic arm, at high inference speeds, without the need for transfer learning. The system is then shown to be capable of performing grasps on an antagonistic physical object dataset benchmark.
\end{abstract}

\section{Introduction}

\noindent In recent years, machine learning has played a key role in determining robotic grasp plan policies for both known and unknown objects \cite{Caldera2018, Du2020}. Broadly speaking, these empirical, or data-driven, implementations of deep neural networks for grasping can be classified into two distinct methods: whether the grasp configurations are sampled and ranked by the network (discriminative models), or directly generated as the output (generative models) \cite{kleeberger2020}. 

Discriminative models rank grasps during execution time and then choose the grasp with the highest score. This can result in carefully evaluated grasps since many grasp poses can be evaluated. However, this can result in higher operational costs due to higher inference times because they require multiple forward passes through the network to consider all available grasps \cite{Mahler2018, Mahler2019, Satish2019b}. Generative models on the other hand directly output a grasp for a whole scene whilst only requiring one forward pass. This allows for rapid closed-loop real time grasp detection, which can be updated more frequently than their discriminative counterparts and can generate multiple grasps per image \cite{Morrison2020, Kumra2020}.

Several exemplars of these generative grasping models rely on the production of simplified planar representations in order to produce a grasp at every point in a scene. By restricting a gripper to move only in two dimensions $(x,y)$, with a corresponding rotation for the gripper around the $z$ axis $(\Theta)$, it is possible to significantly improve the operating speed and training time \cite{Jiang2011, Lenz2015}. These networks are therefore analogous to object detection tasks in computer vision with an added term for gripper orientation \cite{Redmon2015}.

The initial generative grasp convolutional neural network (GG-CNN) by Morrison et al. \cite{Morrison2018} introduced the idea of representing network outputs in the form of \textit{grasp maps}. These outputs decompose a grasp into pixel-wise representations which can be reconstructed at test time, constituting an estimated grasp quality, rotation, and gripper width, to allow for faster training. From the grasp map, the best grasp can be extracted in post-processing in form of the common \textit{rectangle representation} \cite{Jiang2011, Lenz2015}.

\begin{figure}[t]
    \centering
    \includegraphics[scale=0.8]{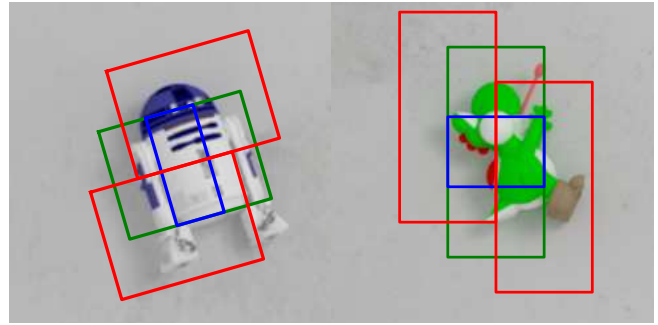}
    \caption{Current generative models for robotic grasping assume a binary representation of grasp labelling. Models are trained to recognise that any grasp centred on a pixel that falls within the centre third (blue) of a correct grasp rectangle (green) are suitable. However, grasps centred on the pixels closer to the edge of the rectangle are less reliable and result in collisions due to incorrect labelling (shown in red).}
    \label{fig:Overview}
\end{figure}

Subsequent generative models such as the generative residual network (GR-ConvNet) \cite{Kumra2020} or orientation attentive grasp synthesis framework (ORANGE) \cite{Chalvatzaki2020a} have built on these grasp maps to achieve state-of-the-art performance on common large-scale grasp datasets. However, these examples rely on the same binary ground truth generation during training which implements a heuristic that assumes any grasp centred within the middle third (and approximately close angle) of an annotated successful grasp is valid. This is incorrect: as Fig. \ref{fig:Overview} shows how grasps centred on pixels towards the edge of a grasp rectangle can lead to gripper collisions when applied to a robotic arm. Despite high performance according to the commonly accepted intersection over union (IoU) threshold, this likely ignores scenarios which would lead to unsuccessful grasp real world grasp attempts.

In order to address this, we present a modified ground truth to train common generative grasping networks and argue that this approach more closely resembles that of the training data. This is more likely to generate successful grasps plans which better capture the geometrical properties of the grasp pose and avoid potential collisions between the gripper and the grasped object. Therefore, this paper makes the following contributions:
\begin{itemize}
    \item A series of common generative grasping models are trained and compared using a  binary and a Gaussian ground truth map. When using the Gaussian ground truth, we demonstrate the network better generalises to unseen objects on the Jacquard grasping dataset \cite{Depierre2018};
    \item Grasping success determined using the offline \emph{rectangle metric} \cite{Jiang2011} is compared with simulated grasp trials (SGT) to show that these offline heuristics, often presented as the measure of model success, are insufficient on their own to predict real-world grasp performance;
    \item We demonstrate that the model trained on simulated data is capable of direct deployment to a physical robotic arm. When tested on a previously unseen physical object dataset, without any transfer learning, we achieve a high grasp success rate with an inference time of 12-14ms using our lightweight model and 25-28ms with our best performing model.
\end{itemize}


\section{Related Work}




\noindent One of the first examples for generative image-based grasping networks included Lenz \textit{et al.} \cite{Lenz2015}. They showed that generative models could be used for real-world robotic grasping tasks after being trained on the Cornell grasping dataset (CGD) \cite{Jiang2011,Lenz2015}, a small dataset containing 885 images annotated with 8019 correct grasps. This work used a two stage process, one for generation the other for ranking of grasps. This process was rather slow as it had to rank all the grasps for one image using multiple forward passes through the model. The work defined a grasp using a five dimensional representation which outputs a grasp in the form of a rectangle with a position, orientation, and size $(x,y,\theta,h,w)$. It was also noteworthy as it normalised the \textit{rectangle metric}, which accepts grasps that overlap sufficiently with annotated grasps, that is generally used to measure grasping model performance on the CGD. Later work from Redmon and Angelova \cite{Redmon2015} improved the speed in which grasps were proposed using another generative system called SingleGrasp. This introduced a single-stage regression-based neural network for robotic grasping which achieved 88\% accuracy on the CGD. 

More recently, neural networks perform near perfectly on the CGD with Park \textit{et al.} \cite{Park2020} achieving 98.6\% accuracy using a single multi-task neural network that uses relationship reasoning among objects. However, it is difficult to generalise results beyond the CGD on such a small sample of objects, and therefore results are typically provided alongside robotic arm data, either simulated \cite{James2019} or with a real robot arm \cite{Kumra2017}. However, it is common for each study to set their own benchmark for reporting results with variations between robotic arms and lists of standardised objects.

Larger datasets have since been developed to train and help models to generalise to unknown objects, including the Jacquard grasping dataset (JGD) \cite{Depierre2018} that features a set of 54k images and 1.1M annotated correct grasps. One major advantage of this dataset is that a simulated robot arm is provided on-line that allows performance to be tested in the same conditions as the data was generated for a more standardised benchmark, known as the simulated grasp trial (SGT) score. Despite this, for speed and convenience most authors continue to use the rectangle metric to evaluate performance.

Using the larger datasets for training, further improvements to the regression-based neural networks were developed. These include the Generative Grasping Convolutional Neural Network (GG-CNN) by Morrison et al. \cite{Morrison2018} that showed how multiple grasps could be generated simultaneously from a scene by outputting a grasp in the form of \textit{grasp maps}: a pixel-wise image-based representation of a grasp consisting of, for each pixel, a grasp quality score, angle, and gripper width. They obtained a 78\% accuracy on the Jacquard test set, and later 84\% with the slightly larger GG-CNN2 network \cite{Morrison2020}. Performance was further increased by models such as the Generative-Residual convolutional network (GR-ConvNet) \cite{Kumra2020} which used a larger network with residual layers to achieve 94.6\% accuracy on the JGD.

Models that train using these grasp maps do not use raw data to train but instead generate a ground truth that works with multiple outputs. Despite the possibility of multiple annotated correct grasps centred at a given pixel, the grasp map ground truth used in training only contains one angle and width value. Depending on the order the labels are used, there may exist discontinuities in angle and width parameters, which makes task learning more difficult. Chalvatzaki \textit{et al.} \cite{Chalvatzaki2020a} showed that the order in which the grasps overlapped when the ground truth grasp maps were generated affected the model performance. They proposed a change to the output whereby the GG-CNN and widely used UNet model \cite{Ronneberger2015} had their outputs modified to identify multiple discrete grasp orientations. 

Adding an attention mechanism to such generative models further improves performance \cite{Chalvatzaki2020a, Prew2020}. By reducing the emphasis of learning angle and width values in the background of a given image using an attentional loss function, accuracy according to the IoU measure increases substantially, speeding up training time without impacting inference time \cite{Prew2020}.

Furthermore, these approaches use the same binary ground truth quality maps containing the issues illustrated in Fig.~\ref{fig:Overview}, with the exception of ORANGE \cite{Chalvatzaki2020a} which applied a \textit{soft quality map} that slightly reduced the ground truth values away from the centre of grasps. Here, we explore the effect of applying a full Gaussian filter to the quality map, further focusing attention on the best grasp positions. We obtain a substantial improvement in performance in SGT score, i.e. within the simulated physics environment and also demonstrate that the IoU measure of performance does not correlate well with simulated performance.

\section{Grasping Problem}

\noindent The grasping problem we aim to address is that of \cite{Kumra2020, Morrison2020, Chalvatzaki2020a}. The challenge is to take an input image, in our case an RGB-D input $\mathbf{I} = \mathbb{R}^{4 \times h \times w}$, with height $h$ and width $w$ of $320 \times 320$ pixels, and find an optimal grasp configuration:
\begin{equation}
    G_i = (x, y, \Theta_i, W_i)
\end{equation}
where  $(x, y)$ is the centre of the proposed grasp in image pixels, $\Theta_i$ the rotation of the proposed grasp, and $W_i$ is the required gripper width, represented in the image frame of reference $i$. 

This 2D grasp can then be converted to a 3D grasp in real-world coordinates. To execute a grasp proposal in the real world from a grasp rectangle given in image coordinates, the grasp must undergo a series of known transforms:
\begin{equation} \label{eq:transform}
    G_r = t_{RC}(t_{CI}(G_i))
\end{equation}
where $t_{CI}$ is the transform from the 2D image coordinates into the 3D camera frame using known camera intrinsics, and $t_{RC}$ is the transform from the camera frame to the world or robot frame. The grasp pose in the robot frame of reference is then represented as follows:
\begin{equation}
    G_r = (\mathbf{P}, \Theta_r, W_r)
\end{equation}
with $\mathbf{P} = (x, y, z)$ being the centre of the parallel gripper jaws, $\Theta_r$ is the angle of the parallel gripper around the $z$-axis, and $W_r$ is the required width of the tool in mm.

\section{Methodology}

\noindent In this section, we outline the training methodology for our experiments. In Subsection \ref{Dataset}: we summarise the Jacquard dataset which was used for training and testing; in Subsection \ref{Networks} we describe the generative grasping models that are used for training; Subsection \ref{Gaussian} describes our novel contribution in which we alter the ground truth from the Jacquard dataset for training the models using a Gaussian map; and finally Subsection \ref{Training} describes the methods and the loss functions used to train the model.

\subsection{Jacquard Grasping Dataset} \label{Dataset}
\noindent All models are trained on the Jacquard grasping dataset \cite{Depierre2018}, a simulated 2D planar dataset containing 54,485 images of over 11,000 different 3D objects on uniform white backgrounds. Grasps are attempted at many positions, angles, and widths in a simulated physics environment. The images are annotated with over 1.1 million successful grasps, including successful grasps at multiple angles and jaw sizes centred on a given pixel. Unsuccessful grasps are not recorded and highly similar grasps are filtered out so are therefore also not included in the dataset. Every object in the dataset has at least four viewing angles and each viewpoint consists of a single RGB image as well as a perfect depth image recorded from the simulated data and a generated stereo depth image. Only the true simulated depth image was used to train these models.

Performance on the Jacquard dataset can be measured using one of two methods: the intersection over union (IoU), also known as the rectangle metric, and simulated grasp trials (SGT) using the Jacquard server\footnote{Available at: https://jacquard.liris.cnrs.fr/} featuring a simulated arm. Using the rectangle metric, a grasp was considered to be correct if:
\begin{itemize}
    \item the predicted grasp rectangle and a corresponding ground truth grasp rectangle share an intersection over union (IoU) score of greater than 25\%, and
    \item the offset of the predicted grasp rectangle aligns within $30^{\circ}$ with the corresponding ground truth grasp rectangle.
\end{itemize}
Based on Jiang et al. \cite{Jiang2011}, Lenz \textit{et al.} \cite{Lenz2015} reduced the threshold for a grasp to be considered successful from 50\% to 25\%, arguing ``\emph{since a ground truth rectangle can define a large space of graspable rectangles (e.g. covering the entire length of a pen), we consider a prediction to be correct if it scores at least 25\% by this metric}''. The threshold of 25\% has been used to report performance in subsequent studies. 

The IoU (rectangle) method is a fast offline method for assessing model performance as it can be evaluated locally. However, this can lead to inaccuracies as a proposed grasp can meet the criteria for the rectangle metric, but could cause a gripper to collide with or miss an object \cite{Depierre2020a}. The red rectangles shown in Fig.~\ref{fig:Overview} represent grasps that fail to pick up the objects in simulation, but have IoU scores of over 25\% so would be reported as correct in most studies. 

The SGT measure of performance is a more robust metric, conducted on the Jacquard simulation server that performs the proposed grasp in the same simulated environment with the same arm as the data was  collected \cite{Depierre2018}. However, this is more costly in time and computation than the IoU metric. Therefore, we have identified the best performing models, according to the IoU metric, and submitted them to the Jacquard on-line server in order to obtain a more accurate comparison between models using SGT as this is designed to be a more accurate benchmark for evaluating robotic grasping performance. 

\subsection{Generative Grasping Networks} \label{Networks}
\begin{figure}
    \centering
    \includegraphics[width=\columnwidth]{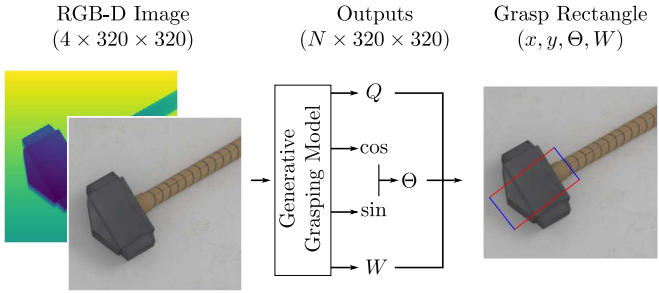}
    \caption{Given an RGB-D ($4\times320\times320$) image, each generative grasping model outputs four \emph{grasp maps}: $Q$, $\cos$, $\sin$, and $W$ which are the same size as the input $N\times320\times320$ with $N$ representing the number of output bins. $\Theta$ is calculated during post processing from $\cos$ and $\sin$ to form the proposed 2-D grasp rectangle. This is formed by taking the max grasp quality pixel score from $Q$ to form the grasp centre $(x,y)$ and the corresponding pixel values from the angle $\Theta$ and width $W$ bin to create a grasp rectangle of $(x,y,\Theta,W)$.}
    \label{fig:Pipeline}
\end{figure}

\noindent The approaches considered in this study are regarded as generative grasping models in that they only require a single pass to generate a grasp proposal. Each network outputs four \textit{grasp maps} which contain a value for each pixel representing different grasping rectangle components: $Q, \Theta^{\cos}, \Theta^{\sin},$ and $W$ (see Fig.~\ref{fig:Pipeline}).

The value $Q$ for each pixel represents the probability of a successful grasp being made centred at the location of the given pixel, and a grasp rectangle can be constructed by taking the corresponding pixel value in the appropriate grasp map with a gripper at angle $\Theta$ and width $W$,  providing the overall image frame of reference output:
\begin{equation}
    \mathbf{G} = (\mathbf{Q}, \Theta^{\cos}, \Theta^{\sin}, \mathbf{W})^{h \times w}
\end{equation}
\begin{itemize}
    \item $\mathbf{Q} \in \mathbb{R}^{h \times w}$ represents a quality map where each pixel is a scalar in the range of 0 to 1, with values nearer to 1 predicting a higher chance of a successful grasp.
    \item $\Theta \in \mathbb{R}^{h \times w}$ is the corresponding angle of the gripper required around the z-axis to grasp an object in the scene and is a value in the range of $[-\frac{\pi}{2}, \frac{\pi}{2}]$ for each pixel. The angle $\Theta$ may be inferred from the network outputs: $\Theta^{\cos}$ and $\Theta^{\sin}$ which are the two decomposed unit vectors of $\Theta$. $\Theta^{\sin}$ is in the range of $[0,1]$ and $\Theta^{\cos}$ in the range of $[-1,1]$. This removes any discontinuities where the angle wraps around $\pm\frac{\pi}{2}$, and provides unique values within $\Theta \in [-\frac{\pi}{2}, \frac{\pi}{2}]$ \cite{ Morrison2018,Hara2017}. The angle of the proposed grasp can be calculated pixel-wise in post-processing by $\Theta = \arctan(\frac{\sin(2\Theta^{sin})}{\cos(2\Theta^{cos})})/2$.
    \item $\mathbf{W} \in \mathbb{R}^{h \times w}$ is the width of the gripper in pixels in the range of $[0, W_{max}]$ which can be converted into real world units using known measurements. $W_{max}$ is the maximum width of the parallel gripper.
\end{itemize}

The output is therefore a grasp proposal for every pixel, along with a quality estimate. To extract a grasp proposal, we take the centre of the rectangle as the pixel position giving maximum $Q$ value and use the corresponding angle $\Theta$ and width $W$ from the same pixel position.

This study trains a variety of generative grasping deep neural network architectures such as the Generative Grasping CNN (GG-CNN2) \cite{Morrison2019}, Generative Residual ConvNet (GR-ConvNet) \cite{Kumra2020}, and the image detection model UNet \cite{Ronneberger2015} according to \cite{Chalvatzaki2020a}. All models are trained using $320 \times 320$ 4-channel RGB-D images. Input data is cropped, resized, and normalised before being processed by the network to match the training data used and depth data is inpainted \cite{Morrison2018, Xue2017}.

Typically when these models are trained, the corresponding ground truth $\Theta^{\cos}$, $\Theta^{\sin}$, and $W$ grasp maps contain pixel values where the angle and width are equivalent to those of a corresponding successful grasp centred at the pixel position in grasp map $Q$. However, due to the structure of the Jacquard dataset, an image contains multiple grasps centred on the same pixel where a variety of gripper angles and widths for a given centred grasp are valid. When using a single grasp map, an arbitrary selection of which angle and width to use at such pixels must be made. The way this choice is made has been shown to affect model performance \cite{Chalvatzaki2020a}. In order to reduce overlapping labelled grasps we employ a technique from the orientation attentive grasp synthesis model (ORANGE) \cite{Chalvatzaki2020a} that separates the grasp angles into $N$ bins with each bin containing a range of $180/N$ degrees. The network then outputs grasp maps for each bin, which allows the network to learn $N$ grasps at each pixel. The output of the network therefore becomes:
\begin{equation}
    \mathbf{G} = (\mathbf{Q}, \Theta^{\cos}, \Theta^{\sin}, \mathbf{W}) ^{N\times h\times w}
\end{equation}
where each of the $N$ dimensions gives the grasp maps restricted to that bin of angles. We compare the models that output grasps as a single bin and when split into 3-bins. In this instance, to reconstruct a grasping rectangle for testing, the maximum $Q$ value across all three bins is taken as the $(x, y)$ grasp centre, with the corresponding $\Theta^{\sin}$, $\Theta^{\cos}$, and $W$ pixel values from the corresponding bin make up the final components of the grasping rectangle. For remaining overlaps, the grasp with the smallest width was used to generate the ground truth, in the same way as \cite{Chalvatzaki2020a}.

One benefit of these generative networks is that a corresponding grasp score is generated at each pixel of an image. In this work, whilst we only consider scenes with single objects, when deployed in scenes with multiple objects, grasp proposals for all objects are generated in a single pass \cite{Kumra2020}.

\subsection{Gaussian Ground Truth Grasp Maps} \label{Gaussian}
\begin{figure}
    \centering
    \includegraphics[width=\columnwidth]{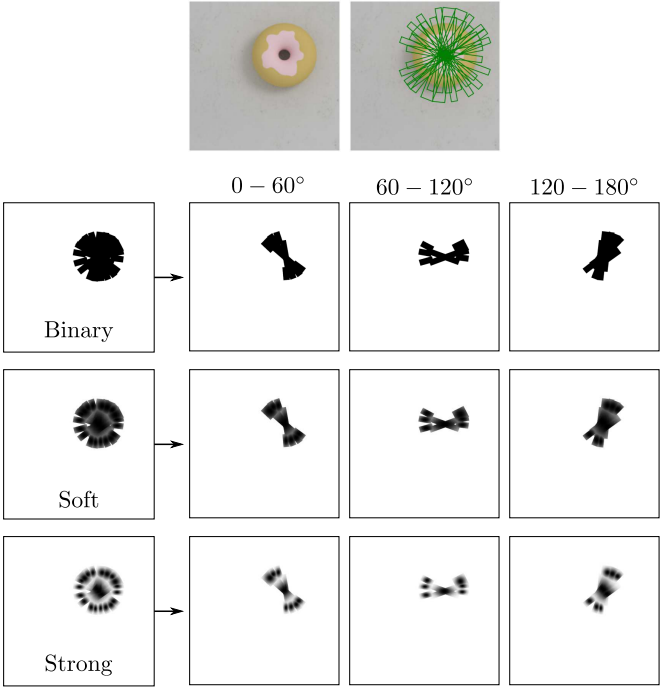}
    \caption{Annotated grasps from the dataset are transformed into ground truth quality maps. All the given grasps for an object are transformed into an image for training. If each picel in the centre third is identified as a suitable grasp centre then the output is as given in the top left. With a Gaussian representation, the ground truth becomes more nuanced, which narrows down the appropriate grasp centres so an end-effector collision is less likely. This ground truth is then separated into 3 buckets so the network is trained to predict the grasp quality score for an associated range of grasp angles.}
    \label{fig:Dataset}
\end{figure}
\noindent An ideal ground truth would be generated by using a physics engine to simulate a grasp at each possible angle at each pixel location, assigning the value 1 if there is a successful grasp at some angle at that location. However this would require tens of millions of simulations per input and is therefore computationally infeasible on large datasets. 

We must use some method to infer grasp quality values at points that have not been directly simulated. Previous versions of generative grasping networks such as GGCNN2 \cite{Morrison2019} and GR-ConvNet \cite{Kumra2020} have trained networks using grasp maps where ground truth values for $Q$ are represented as a binary image mask (see Fig. \ref{fig:Dataset}). The traditional binary $Q$ grasp map assumes that all pixels within the centre third of a grasp rectangle are correct grasps and assigns a ground truth $\hat{Q}$ (where $\hat{}$ represents the associated ground truth grasp map) value of $1$ if a pixel falls within this section of any grasping rectangle and $0$ otherwise. As previously discussed, this heuristic for generating ground truth values results in inaccuracies such as quality scores that are centred away from the object, as illustrated in Fig~\ref{fig:Overview}.

Therefore, we propose a Gaussian heuristic for generating ground truths and perform experiments comparing this against the binary heuristic. Only the centre pixel of a successful simulated grasp is assigned a quality score $\hat{Q}$ value of 1, and the assigned $\hat{Q}$ value gradually decays to near zero according to a Gaussian distribution. The strength of the Gaussian was selected using the hyperparameter $\sigma$, which alters the how sharply the $\hat{Q}$ value reduces away from the centre. A smaller $\sigma$ value represents a smaller standard deviation focuses attention on the centres of successful simulated grasps, whereas a large $\sigma$ allows the network to generalise successful grasps to similar areas in the surrounding pixels. 

We apply this Gaussian filter in one of two ways against the binary quality map: Firstly, the \textit{soft quality map}, as described alongside the ORANGE model \cite{Chalvatzaki2020a}, and our \textit{strong quality map}. For the soft quality map, a Gaussian filter is applied, however, there remains a minimum floor value on the centre third of the grasping rectangle. The centre of the grasping rectangle has a $\hat{Q}$ value of 1, and decays towards a minimum value (0.9) according to the equation:
\begin{equation}
    \hat{Q}(x,y) = \max_{g}\left\{\min\left\{ \frac{\mathcal{N}(d,\,\sigma^{2})}{\mathcal{N}(0,\,\sigma^{2})} \delta,0.9\delta\right\}\right\}
\end{equation}
where the $\hat{Q}$ maximum is generated over all annotated grasps $g$. $d=d((x,y),g)$ is the distance of the pixel $(x,y)$ from the centre of the grasp $g$. $\delta=\delta((x,y),g)$ is an indicator function taking value 1 if $(x,y)$ is in the centre third of the grasping rectangle of $g$ and value 0 otherwise, and $\sigma$ is the hyperparameter determining the strength of the Gaussian. In this case $\sigma=2$ according to the ORANGE model (Fig.~\ref{fig:Dataset}). This ensures that the network is taught to recognise the centre of the grasping rectangle as a better location for grasp approximation, although, this still considers all the centre third to be valid and therefore results in the same problems as the binary map.

We present an alternative to this method, which is referred to as the \textit{strong quality map}: this removes the minimum filter from the soft quality map, and is defined in the following equation:
\begin{equation}
    \hat{Q}(x,y) = \max_{g}\left\{ \frac{\mathcal{N}(d,\,\sigma^{2})}{\mathcal{N}(0,\,\sigma^{2})}\times \delta\right\}.
\end{equation}
$\sigma$ is varied as an extra hyperparameter to find the optimal distribution of grasp centres. By removing the minimum floor, the aim is to better train the network to recognise appropriate grasps by further distinguishing grasp centres between 0-1.

\subsection{Training Method} \label{Training}
\noindent For training and testing, the dataset is split 90/10\% into each set respectively according to the same methods used by \cite{Kumra2020, Chalvatzaki2020a}, with no data augmentation applied during either stage. This leaves a total of 5449 grasping scenes from the dataset to form the test set. We use the same test set to evaluate both the traditional Intersection over Union (IoU) metric and simulated grasp trial-based (SGT) criterion.

Colour pixel values are normalised to the range of $[0,1]$ before subtraction of the image mean to zero-centre the image data. Depth data is also normalised to the range of $[-1,1]$ before a zero-centre via mean subtraction and subsequent clamping of values within this range. All models are trained using the ADAM optimser \cite{Kingma2015} and early stopping is used once the learning rate plateaus after a number of epochs.

Models are trained with their original loss function as well as the \textit{positional loss function} from \cite{Prew2020}. This new loss provides a lightweight attention mechanism to generative grasping models and is performed by multiplying loss contribution from angle and width values by the $\hat{Q}$ value at that pixel. This does not penalise the network for angle and width errors away from positions of where a successful grasp can occur, focusing attention on errors at successful grasp positions. This means that the GG-CNN2 and UNet models are trained using an MSE loss function and the GR-ConvNet model is trained using smooth L1 loss. Following~\cite{Chalvatzaki2020a}, the losses are also scaled by multiplying them with the number of discretised angle bins $N$ and thus making the overall loss for the network equal to:
\begin{equation}
    \mathcal{L} = N \times \left(L(Q) + L(\Theta^{\cos}) + L(\Theta^{\sin}) + L(W)\right)
\end{equation}
with $\mathcal{L}$ representing the loss for the given network and $L$ representing the individual MSE or smooth L1 loss for the given network. The positional loss $\mathcal{L}_\mathcal{P}$ function is then given by:
\begin{equation}
    \mathbf{\mathcal{L}_{\mathcal{P}}} = N \times \left(L(Q) + \hat{Q}(L(\Theta^{\cos})+L(\Theta^{\sin})+L(W))\right)
\end{equation}

In generating the ground truth, for situations where multiple grasps centred on the same pixel values existed with different corresponding angles and widths: the smallest sized grasp is used \cite{Chalvatzaki2020a}. Similarly, a half jaw size is adopted during testing \cite{Chalvatzaki2020b}. Results from testing are first reported using the IoU (rectangle) metric to establish a quick offline evaluation and the best performing models for each network architecture are sent to the Jacquard server for a robust comparison.

\subsection{Physical Experiments}
\begin{figure}
    \centering
    \includegraphics[width=0.7\columnwidth]{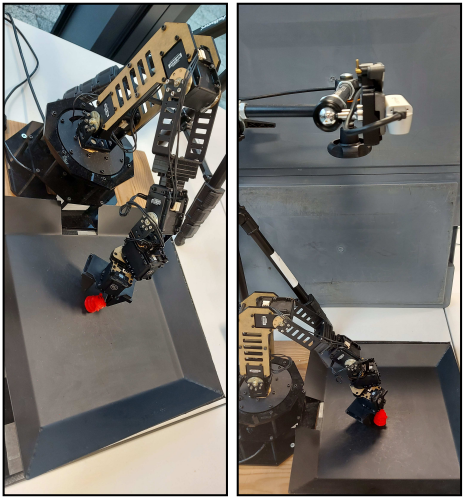}
    \caption{The setup of the WidowX robot arm used in the physical experiments, with the camera positioned above the scene.}
    \label{fig:arm}
\end{figure}

\noindent In addition to the simulated grasp trial data presented, experiments utilising a \textit{WidowX robot arm} are implemented to show that the model can easily transfer to a physical real-world setup. The setup takes an image from above using an Intel RealSense SR300 RGB-D camera, in the same orientation of that used in the JGD, and generates the given grasp proposal from the model for the given object. The robot arm used in this work is a 6 degrees of freedom (6DoF) WidowX arm from Interbotix Labs: a 1DoF rotating base, three 1DoF joints, a 1DoF rotating wrist, and a 1DoF parallel plate gripper with minimum 1cm and maximum 3cm width. The setup is shown in Fig.~\ref{fig:arm} and is the same low-cost arm as used in REPLAB \cite{Yang2019}. Grasp plan motions are created using ROS inverse kinematics and planned with the \textit{MoveIt} package.

Using Equation~\ref{eq:transform}, the 2D output from the model is transformed into the robots frame of reference by taking the maximum pixel coordinate $(x,y)$ from the grasp quality score, and using the corresponding depth coordinate from an RGB-D camera to the depth point in 3D space $z$ to form a 3D grasp location $(x,y,z)$. 

Nowadays, single object grasping in uncluttered scenes is highly accurate. Therefore, to show the model is capable of transferring knowledge to completely unrelated objects, a standardised set of 3D printed objects is used for testing called the evolved grasping analysis dataset (EGAD)\footnote{Available at https://dougsm.github.io/egad/} \cite{Morrison2020}. This features a diverse range of objects of varying difficulty and complexity, including simple and antagonistic examples. The dataset ranges from A0 to G6. Increasing lettering represents more difficult to grasp objects but should represent the similar grasp difficulty whereas increased numbering corresponds to increased complexity.

Data is presented using the model as trained with the simulated Jacquard data with no transfer learning involved. This is to show ease of transferability to other settings and that the model can easily generalise to similar settings. To this effect, the same grasping methodology as used in the original EGAD study \cite{Morrison2020} is repeated. Each object is thrown randomly into the arena and a grasp is attempted 20 times for each object. The grasp is considered successful if the object is lifted and stable above the arena once the gripper has closed. The object is then dropped back randomly into the arena for the next attempt. If the object is unsuccessfully grasped then it is manually reset by throwing it back into the arena randomly, to ensure the network is not continuously attempting incorrect grasps.

\section{Results and Discussion} \label{Results}

\setlength\tabcolsep{4pt}
\begin{table}
\renewcommand{\arraystretch}{1.3}
\caption{Performance on the test portion of the Jacquard grasping dataset according to the IoU metric at the 25\% threshold}
\label{table:Gaussian}
\centering
\begin{tabular}{lcccccccc}
\toprule
\multirow{2}{*}{\bfseries{Model}} & \multirow{2}{*}{\bfseries{Loss}} & \multirow{2}{*}{\bfseries{Bins}} & \bfseries{Binary} & \bfseries{Soft}  & \multicolumn{4}{c}{\bfseries{Strong}}\\
&&&$\sigma$&2 & 2 & 1 & 0.5 & 0.25\\
\hline
\multirow{4}{*}{GG \cite{Morrison2019}} &\multirow{2}{*}{MSE}&1&\textbf{87.87}&87.69&86.79&87.50&86.86&85.74\\
& &3&88.00&\textbf{88.73}&87.83&87.65&86.93&86.02\\
&\multirow{2}{*}{Pos} & 1 &91.21&\textbf{91.99}&88.13&90.18&90.18&89.96\\
& &3&\textbf{93.98}&88.59&91.39&92.90&90.93&92.42\\
\hline
\multirow{4}{*}{GR\cite{Kumra2020}}&\multirow{2}{*}{ SL1}&1&\textbf{90.86}&90.82&90.16&90.77&89.74&89.10\\
&&3&91.65&92.05&91.98&\textbf{92.40}&91.41&\textbf{92.40}\\
&\multirow{2}{*}{Pos}&1&92.27&91.89&91.76&91.47&91.98&\textbf{92.35}\\
&&3&\textbf{93.69}&91.82&93.47&90.99&93.21&92.40\\
\hline
\multirow{4}{*}{UN \cite{Chalvatzaki2020a}} &\multirow{2}{*}{MSE} &1&\textbf{90.55}&89.52&90.48&89.94&89.94&89.91\\
&&3&\textbf{91.78}&91.14&90.51&89.21&89.67&89.89\\
&\multirow{2}{*}{Pos}&1&\textbf{93.61}&93.30&92.62&91.69&92.48&92.18\\
&&3&\textbf{94.66}&93.45&93.98&94.35&93.83&92.59\\
\bottomrule
\end{tabular}
\end{table}

\noindent In this work, a series of generative grasping models including: the generative grasping convolutional neural network (GG-CNN2) \cite{Morrison2019}; generative residual convolutional network (GR-ConvNet) \cite{Kumra2020}; and UNet architectures \cite{Chalvatzaki2020a,Ronneberger2015}, are trained on the Jacquard grasping dataset \cite{Depierre2018}. The results from both the intersection over union (IoU) metric \cite{Jiang2011, Lenz2015} and simulated grasp trials (SGT) are reported for the same unseen data. The IoU metric is used as a quick offline metric for evaluating performance of all models and then the best performing models are tested using the simulated physics environment on the Jacquard test server \cite{Depierre2018}, as this is a more robust evaluation of performance. 

Each model is trained using the traditional binary ground truth grasp map introduced in \cite{Morrison2018}, the \emph{soft quality map} from \cite{Chalvatzaki2020a}, and the \emph{strong quality map}, as described in Subsection~\ref{Gaussian}. The strength of the Gaussian filter $\sigma$ is varied to find the optimal spread of trainable parameters for the dataset. The best performing model is then applied in a real world setting, to show the model is capable of generalising to completely unseen objects using a low-cost robot arm.

\subsection{Offline versus Simulated Performance}
\noindent The overall performance of each model when measured using the typical IoU threshold of 25\% is reported in Table \ref{table:Gaussian}. Firstly, this data shows that models trained with three output angle bins perform better than those limited to one angle bin, as previously shown in \cite{Chalvatzaki2020a, Chalvatzaki2020b}. Similarly, models trained with the \emph{positional loss} function outperform the same model when trained with each respective base loss function, as previously shown in \cite{Prew2020}. As far as we are aware, this is the first work to combine both these methods concurrently. This approach achieves the best reported IoU metric for each model, showing these two improvements complement one other, which has not been previously demonstrated. Therefore, all models compared in the SGT results in Table ~\ref{table:SGT} feature models trained with both methods in unison.

From the IoU measure of 25\%, the conclusion would be that there is little difference when comparing the same models on different ground truth maps. The best reported value overall is achieved by the UNet model with a generated binary ground truth $(94.66\%)$, which slightly outperforms the same model when trained with the strong $(94.35\%)$ Gaussian maps followed by the soft quality ground truth map $(93.45\%)$. This conclusion, however, does not hold when we consider the more robust SGT results below.

In Table~\ref{table:SGT}, when we analyse performance at higher IoU thresholds, the difference between the binary and Gaussian methods becomes more apparent. Despite close results at the traditional 25\% threshold, the grasping performance of the models separates at higher thresholds. An increased grasp success at larger IoU thresholds demonstrate grasps that highly resemble that of the test set. For example, there is a broader separation between the models at the 75\% threshold, showing that models are not learning how to best recreate the training data. In each case, at higher thresholds, the soft Gaussian map shows the greatest decrease in grasp success across all models, whereas the strong Gaussian method remains the most consistent. This shows an inherent issue with the IoU metric as it results in saturated grasp performance, particularly at low threshold values. The average IoU is also included to display performance across all thresholds.

\begin{table}
\renewcommand{\arraystretch}{1.3}
\caption{Performance of models trained with three output bins and positional loss function on the Jacquard grasping dataset.}
\setlength\tabcolsep{5.5pt}
\label{table:SGT}
\centering
\begin{tabular}{llcccccc}
\toprule
\multirow{2}{*}{\bfseries{Model}} & \multirow{2}{*}{\bfseries{Map}} & \multicolumn{5}{c}{\bfseries{IoU} } & \bfseries{SGT}\\
&& \bfseries{25\%} & \bfseries{30\%} & \bfseries{50\%} & \bfseries{75\%} & \textbf{Avg} & \bfseries{\%}\\
\hline
\multirow{3}{*}{GG \cite{Morrison2019}}
&Binary&93.98&92.33&79.61&30.21&62.46&85.41\\ 
&Soft  &88.59&85.15&69.41&25.49&57.40&85.43\\ 
&Strong&92.90&91.14&80.29&39.20&64.13&86.58\\ 
\hline
\multirow{3}{*}{GR \cite{Kumra2020}}
&Binary&93.69&91.83&83.01&39.37&65.57&85.36\\ 
&Soft  &91.15&88.81&79.52&23.47&61.16&83.06\\ 
&Strong&93.21&90.95&82.44&49.62&66.98&85.89\\ 
\hline
\multirow{3}{*}{UN \cite{Chalvatzaki2020a}}
&Binary&\textbf{94.66}&\textbf{93.25}&\textbf{84.42}&43.11&66.61&85.69\\ 
&Soft  &93.45&91.43&82.27&40.03&65.05&85.78\\ 
&Strong&93.98&92.31&83.87&\textbf{50.71}&\textbf{67.69}&\textbf{87.94}\\ 
\bottomrule
\end{tabular}
\end{table}

Since the evaluation of these models in simulation takes significantly longer to produce than the typical offline evaluation, only the highest performing models were evaluated in this manner. This was according to both the fast rectangle metric and the average (IoU-Avg) score. Whilst the effects of altering the hyperparameter $\sigma$ while using strong Gaussian maps are considered, there is no clear optimal value. Generally performance benefits from a moderate value in which the Gaussian map does not include the edges of the grasping rectangle but maintains a large enough collection of high quality grasp centres to learn from. We note that the optimal Gaussian scaling is likely tied to the density of labelled grasps in the dataset for a given object and a given model. As the JGD contains a high density of grasp labels, it is likely that the optimal scaling factor is smaller than for a dataset with more sparsely sampled grasp labels, such as the CGD \cite{Jiang2011,Lenz2015}, and requires fine tuning with other datasets.

Together, these results suggest that models trained with the strong Gaussian map learn to predict grasp rectangles closer to the original simulated grasps, which have previously confirmed a successful grasp on the object. This is represented by the higher IoU-Avg score of proposed grasps as well as confirmation on SGT results, which show predicted grasps are less likely to result in gripper collisions during implementation. Therefore, this difference in SGT performance highlights the inherent problem with only reporting the traditional IoU metric for predicting real-world performance. Here, the strong Gaussian map also achieves the highest accuracy on the more robust measure of performance as this also considers grasps not included in the dataset. The best performing UNet model reports a total of 87.94\% successful grasps on the same 5449 object scenes as used to measure the IoU compared to only 85.69\% using the binary grasp map. This is also over a 2\% performance increase over the highest reported result for the SGT metric so far~\cite{Depierre2020a}. We suggest in future work that if a fast, offline estimate of performance is presented: the average IoU (IoU-Avg) score of proposed grasps should be reported, alongside the commonly used IoU metric at multiple thresholds, as a predictor of robotic arm/SGT success.

\subsection{EGAD Results}
\begin{figure}[t!]
    \centering
    \includegraphics[width=\columnwidth]{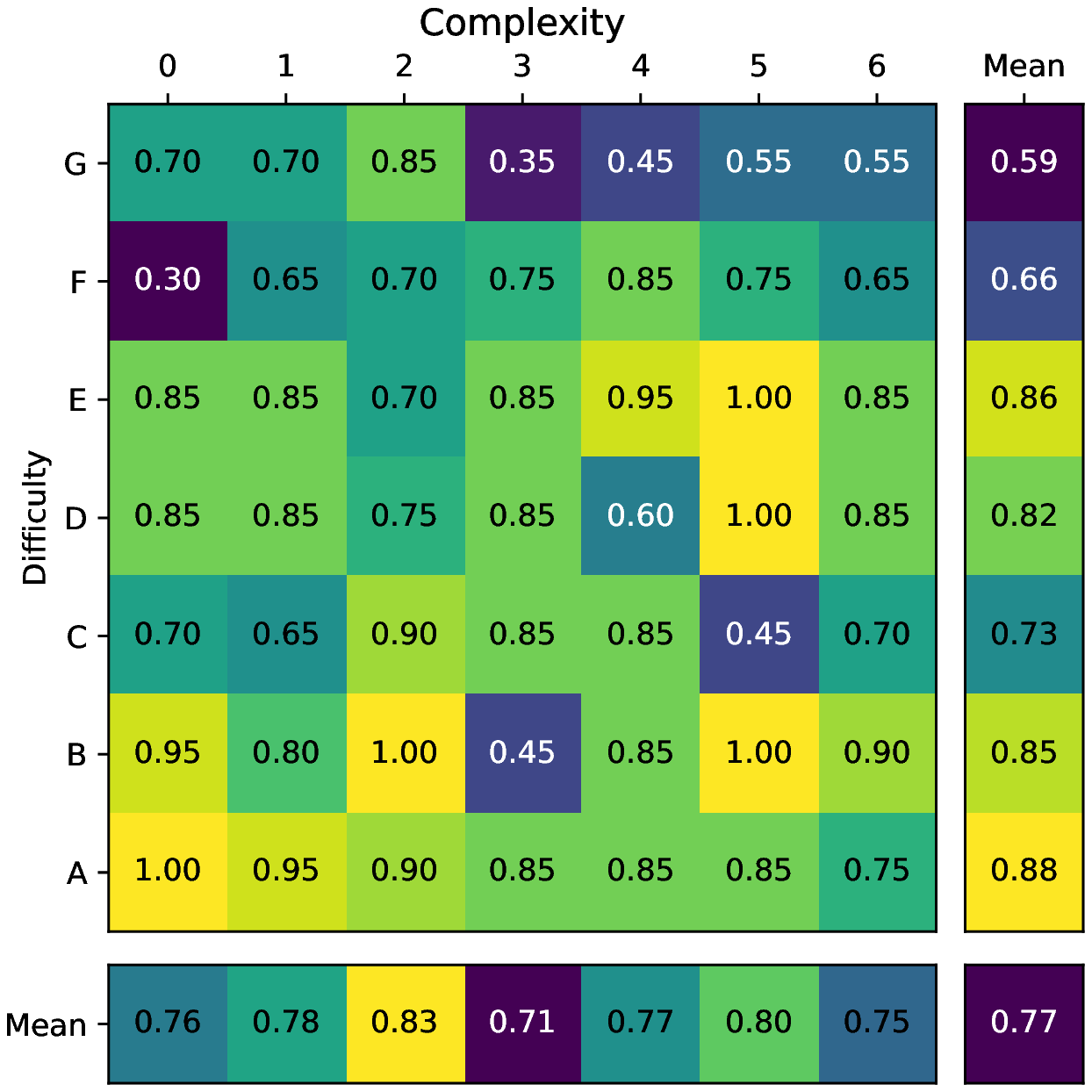}
    \caption{Average grasp success rate for each object in the EGAD \cite{Morrison2020} evaluation dataset. Outer cells show the mean for that row and column.}
    \label{fig:EGAD_results}
\end{figure}

\noindent In addition to the SGT results, which show that the trained model is capable of producing grasps in the environment native to the training procedure, the model was also applied to a standard low-cost WidowX robotic arm. The arm is tasked with picking up each object from the EGAD \cite{Morrison2020} evaluation set 20 times for a total of 980 grasp trials. A grasp is considered successful if a correct plan is made to attempt an object grasp and the arm is able to lift the object above the arena after closure of the gripper. We apply an open-loop grasping method where a grasp plan is made in the same way as the SGT. The camera is placed above the scene to mimic that of the JGD but otherwise no transfer learning took place. The results of these tests, are shown in Fig.~\ref{fig:EGAD_results}.

The applied model performs relatively well overall despite only being trained on simulated data. The model maintains reasonable consistency across all object complexities, and generally decreases in performance as object difficulty increases. The model performs best when grasping the easiest objects (A) and slightly dropping in performance towards the most difficult objects (G). In some trials, the robot is even able to achieve perfect or near perfect results. In all trials, the robot made an accurate attempt to grasp the object in the scene, which shows that the model is able to be applied with high accuracy without transfer learning. This results in an overall mean accuracy of 77\% over all grasps attempted which, while not directly comparable due to the difference in arm setup, is higher than the 58\% accuracy achieved by only the base GG-CNN model in the original study \cite{Morrison2020}.

\begin{figure}
    \centering
    \includegraphics[width=\columnwidth]{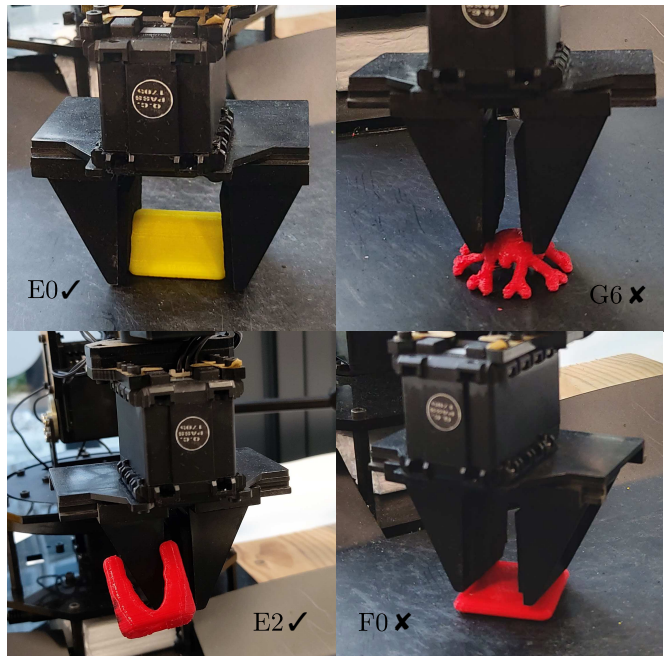}
    \caption{Example grasps on using a low-cost robotic arm with a parallel plate gripper. Most objects are grasped by reaching across the principal axis of the object (top left), however the model is also capable of plan grasps that only reached across parts of the object (bottom left). The most common failure cases are due to object difficulty where the model suggested grasps unsuitable for the type of gripper used (right images).}
    \label{fig:EGADexamples}
\end{figure}

Whilst this robotic implementation requires an external depth camera, very few examples are failures as a result of an incorrect gripper depth. Most failure cases observed are due to designed object difficulty, such as grasping parts of the object with angled sides or raised edges, see Fig.~\ref{fig:EGADexamples} for examples. Other cases where grasp success is low, such as object B3 or D4, resulted from a lack of knowledge about the gripper. The model would predict grasps that resulted in object collisions with the gripper plates by grasping along an unfriendly axis relative to the robot end-effector, cases which would otherwise be fine using a narrower pinch gripper. These could be improved by further training with knowledge of the specific end-effector but the JGD is primarily designed for parallel-jaw grippers like the one used here.

\begin{table}
    \renewcommand{\arraystretch}{1.3}
    \caption{Inference time for a single grasp on grasping models.}
    \centering
    \begin{tabular}{l r r} \toprule
    Model & Model Parameters & Inference Time (ms)\\
    \hline
    Google Grasp \cite{Levine2018} & 1M & 200-500\\
    GQ-CNN \cite{Mahler2018} & 18M & 800\\
    FC-GQ-CNN \cite{Satish2019b} & - & 625\\
    \hline
    GGCNN2 \cite{Morrison2019} & 72k &  12-14\\
    GR-ConvNet \cite{Kumra2020} & 1.9M & 38-40\\
    UNet \cite{Ronneberger2015} & 14.7M & 25-28\\
    \bottomrule
    \end{tabular}
    \label{tab:Inference Time}
\end{table}

Without transferring to the new scene the model still achieves a relatively high success rate. However, it is noted that this is a simple task with only one object per scene. While other networks in more recent studies can achieve high success rate by gripping in a 3-D space (e.g. \cite{Satish2019b}), these lightweight generative models balance high accuracy with much faster grasp detection speed. As a result, they can operate at greater speed than other discriminative models, as shown in Table.~\ref{tab:Inference Time}. The inference times for our models are collected using GPU-acceleration on a NVidia GTX 1080Ti graphics card in \emph{PyTorch} 1.3 with CUDA 11, taking the minimum and maximum inference speeds over the test set.

Future generative grasping models may benefit from an inherent depth module to directly predict $(x,y,z)$ grasps without the need for the external transformation. Performance would likely be improved using more accurate robotic grippers and transfer to specific scenes or grippers, e.g. \cite{Kleeberger2020a}. However, this work intends to minimise extensive retraining to reinforce the generalisability of the model.

\section{Conclusion}

\noindent We evaluate the effect of training common generative robotic grasping models using an applied Gaussian filter on a modified ground truth representation. These results show that using both the attentional \emph{positional loss function}, in addition to discrete orientation-specific outputs, together improves grasping performance with little to no overhead. 

Furthermore, the traditional rectangle metric, is insufficient for predicting grasp success on robotic arms. These experiments show that models trained using a Gaussian ground truth, whilst showing negligible performance difference on the rectangle metric, were better able to propose appropriate grasps when testing on a simulated robot arm. Our best model achieves 87.94\% grasp success according to the SGT, which is $>2\%$ performance increase over the previous state of the art on this benchmark \cite{Depierre2020a}. Therefore, we suggest the addition of the IoU-Avg score as an offline metric for predicting real-world model performance. 

This data is further supplemented with real-world data to show the model is capable of transferring to a previously unseen physical object dataset. The trained model achieves high performance even on complex and difficult to grasp objects. Therefore, we reinforce the need for testing of models on physical benchmarks in addition to offline measures.

\section*{Acknowledgements}
\noindent This work was funded by UKRI EPSRC. For the purpose of open access, the authors have applied a Creative Commons Attribution (CC BY) license to the Accepted Manuscript version arising.



\bibliographystyle{IEEEtran}
%
\bibliography{root.bbl}

\begin{thebibliography}{10}
\providecommand{\url}[1]{#1}
\csname url@samestyle\endcsname
\providecommand{\newblock}{\relax}
\providecommand{\bibinfo}[2]{#2}
\providecommand{\BIBentrySTDinterwordspacing}{\spaceskip=0pt\relax}
\providecommand{\BIBentryALTinterwordstretchfactor}{4}
\providecommand{\BIBentryALTinterwordspacing}{\spaceskip=\fontdimen2\font plus
\BIBentryALTinterwordstretchfactor\fontdimen3\font minus
  \fontdimen4\font\relax}
\providecommand{\BIBforeignlanguage}[2]{{%
\expandafter\ifx\csname l@#1\endcsname\relax
\typeout{** WARNING: IEEEtran.bst: No hyphenation pattern has been}%
\typeout{** loaded for the language `#1'. Using the pattern for}%
\typeout{** the default language instead.}%
\else
\language=\csname l@#1\endcsname
\fi
#2}}
\providecommand{\BIBdecl}{\relax}
\BIBdecl

\bibitem{Caldera2018}
S.~Caldera, A.~Rassau, and D.~Chai, ``{Review of Deep Learning Methods in
  Robotic Grasp Detection},'' \emph{Multimodal Technologies and Interaction},
  vol.~2, no.~3, pp. 1--24, sep 2018.

\bibitem{Du2020}
G.~Du, K.~Wang, S.~Lian, and K.~Zhao, ``{Vision-based robotic grasping from
  object localization, object pose estimation to grasp estimation for parallel
  grippers: a review},'' \emph{Artificial Intelligence Review}, 2020.

\bibitem{kleeberger2020}
K.~Kleeberger, R.~Bormann, W.~Kraus, and M.~F. Huber, ``{A Survey on
  Learning-Based Robotic Grasping},'' \emph{Current Robotics Reports}, vol.~1,
  no.~4, pp. 239--249, dec 2020.

\bibitem{Mahler2018}
\BIBentryALTinterwordspacing
J.~Mahler, M.~Matl, X.~Liu, A.~Li, D.~Gealy, and K.~Goldberg, ``{Dex-Net 3.0:
  Computing Robust Vacuum Suction Grasp Targets in Point Clouds Using a New
  Analytic Model and Deep Learning},'' in \emph{IEEE International Conference
  on Robotics and Automation}.\hskip 1em plus 0.5em minus 0.4em\relax IEEE, sep
  2018, pp. 5620--5627. [Online]. Available:
  \url{http://arxiv.org/abs/1709.06670}
\BIBentrySTDinterwordspacing

\bibitem{Mahler2019}
J.~Mahler, M.~Matl, V.~Satish, M.~Danielczuk, B.~DeRose, S.~McKinley, and
  K.~Goldberg, ``{Learning ambidextrous robot grasping policies},''
  \emph{Science Robotics}, vol.~4, no.~26, jan 2019.

\bibitem{Satish2019b}
V.~Satish, J.~Mahler, and K.~Goldberg, ``{On-policy dataset synthesis for
  learning robot grasping policies using fully convolutional deep networks},''
  \emph{IEEE Robotics and Automation Letters}, vol.~4, no.~2, pp. 1357--1364,
  2019.

\bibitem{Morrison2020}
\BIBentryALTinterwordspacing
D.~Morrison, P.~Corke, J.~Leitner, and J.~Leitner, ``{EGAD! An Evolved Grasping
  Analysis Dataset for Diversity and Reproducibility in Robotic
  Manipulation},'' \emph{IEEE Robotics and Automation Letters}, vol.~5, no.~3,
  pp. 4368--4375, mar 2020. [Online]. Available:
  \url{http://arxiv.org/abs/2003.01314}
\BIBentrySTDinterwordspacing

\bibitem{Kumra2020}
\BIBentryALTinterwordspacing
S.~Kumra, S.~Joshi, and F.~Sahin, ``{Antipodal Robotic Grasping using
  Generative Residual Convolutional Neural Network},'' pp. 1--8, sep 2020.
  [Online]. Available: \url{http://arxiv.org/abs/1909.04810}
\BIBentrySTDinterwordspacing

\bibitem{Jiang2011}
Y.~Jiang, S.~Moseson, and A.~Saxena, ``{Efficient grasping from RGBD images:
  Learning using a new rectangle representation},'' in \emph{International
  Conference on Robotics and Automation}.\hskip 1em plus 0.5em minus
  0.4em\relax IEEE, 2011, pp. 3304--3311.

\bibitem{Lenz2015}
I.~Lenz, H.~Lee, and A.~Saxena, ``{Deep learning for detecting robotic
  grasps},'' \emph{International Journal of Robotics Research}, vol.~34, no.
  4-5, pp. 705--724, 2015.

\bibitem{Redmon2015}
J.~Redmon and A.~Angelova, ``{Real-time grasp detection using convolutional
  neural networks},'' in \emph{IEEE International Conference on Robotics and
  Automation}, 2015, pp. 1316--1322.

\bibitem{Morrison2018}
\BIBentryALTinterwordspacing
D.~Morrison, P.~Corke, and J.~Leitner, ``{Closing the Loop for Robotic
  Grasping: A Real-time, Generative Grasp Synthesis Approach},'' pp. 1--10,
  2018. [Online]. Available: \url{http://arxiv.org/abs/1804.05172}
\BIBentrySTDinterwordspacing

\bibitem{Chalvatzaki2020a}
G.~Chalvatzaki, P.~Maragos, J.~Peters, and N.~Gkanatsios, ``{Revisiting Grasp
  Map Representation with a Focus on Orientation in Grasp Synthesis},''
  \emph{ArXiv}, 2020.

\bibitem{Depierre2018}
A.~Depierre, E.~Dellandrea, and L.~Chen, ``{Jacquard: A Large Scale Dataset for
  Robotic Grasp Detection},'' in \emph{IEEE International Conference on
  Intelligent Robots and Systems}, 2018, pp. 3511--3516.

\bibitem{Park2020}
D.~Park, Y.~Seo, D.~Shin, J.~Choi, and S.~Y. Chun, ``{A Single Multi-Task Deep
  Neural Network with Post-Processing for Object Detection with Reasoning and
  Robotic Grasp Detection},'' in \emph{IEEE International Conference on
  Robotics and Automation}, 2020, pp. 7300--7306.

\bibitem{James2019}
\BIBentryALTinterwordspacing
S.~James, P.~Wohlhart, M.~Kalakrishnan, D.~Kalashnikov, A.~Irpan, J.~Ibarz,
  S.~Levine, R.~Hadsell, and K.~Bousmalis, ``{Sim-to-Real via Sim-to-Sim:
  Data-efficient Robotic Grasping via Randomized-to-Canonical Adaptation
  Networks},'' in \emph{IEEE Conference on Computer Vision and Pattern
  Recognition}, 2019. [Online]. Available:
  \url{https://arxiv.org/pdf/1812.07252.pdf}
\BIBentrySTDinterwordspacing

\bibitem{Kumra2017}
S.~Kumra and C.~Kanan, ``{Robotic grasp detection using deep convolutional
  neural networks},'' \emph{IEEE International Conference on Intelligent Robots
  and Systems}, vol. 2017-Septe, pp. 769--776, 2017.

\bibitem{Ronneberger2015}
O.~Ronneberger, P.~Fischer, and T.~Brox, ``{U-Net: Convolutional Networks for
  Biomedical Image Segmentation},'' in \emph{International Conference on
  Medical image computing and computer-assisted intervention}, 2015, pp.
  234--241.

\bibitem{Prew2020}
\BIBentryALTinterwordspacing
W.~Prew, T.~Breckon, M.~Bordewich, and U.~Beierholm, ``{Improving Robotic
  Grasping on Monocular Images Via Multi-Task Learning and Positional Loss},''
  \emph{International Conference on Pattern Recognition}, 2020. [Online].
  Available: \url{http://arxiv.org/abs/2011.02888}
\BIBentrySTDinterwordspacing

\bibitem{Depierre2020a}
\BIBentryALTinterwordspacing
A.~Depierre, E.~Dellandr{\'{e}}a, and L.~Chen, ``{Scoring Graspability based on
  Grasp Regression for Better Grasp Prediction},'' \emph{arXiv}, 2020.
  [Online]. Available: \url{http://arxiv.org/abs/2002.00872}
\BIBentrySTDinterwordspacing

\bibitem{Hara2017}
\BIBentryALTinterwordspacing
K.~Hara, R.~Vemulapalli, and R.~Chellappa, ``{Designing Deep Convolutional
  Neural Networks for Continuous Object Orientation Estimation},'' pp. 1--10,
  2017. [Online]. Available: \url{http://arxiv.org/abs/1702.01499}
\BIBentrySTDinterwordspacing

\bibitem{Morrison2019}
D.~Morrison, P.~Corke, and J.~Leitner, ``{Learning robust, real-time, reactive
  robotic grasping},'' \emph{International Journal of Robotics Research},
  vol.~39, no. 2-3, pp. 183--201, mar 2019.

\bibitem{Xue2017}
H.~Xue, S.~Zhang, and D.~Cai, ``{Depth Image Inpainting: Improving Low Rank
  Matrix Completion with Low Gradient Regularization},'' \emph{IEEE
  Transactions on Image Processing}, vol.~26, no.~9, pp. 4311--4320, 2017.

\bibitem{Kingma2015}
D.~Kingma and J.~L. Ba, ``{Adam: A method for stochastic optimization},'' in
  \emph{International Conference on Learning Representations}, 2015.

\bibitem{Chalvatzaki2020b}
\BIBentryALTinterwordspacing
G.~Chalvatzaki, N.~Gkanatsios, P.~Maragos, and J.~Peters, ``{Orientation
  Attentive Robot Grasp Synthesis},'' 2020. [Online]. Available:
  \url{http://arxiv.org/abs/2006.05123}
\BIBentrySTDinterwordspacing

\bibitem{Yang2019}
\BIBentryALTinterwordspacing
B.~Yang, D.~Jayaraman, J.~Zhang, and S.~Levine, ``{REPLAB: A reproducible
  low-cost arm benchmark for robotic learning},'' in \emph{Proceedings - IEEE
  International Conference on Robotics and Automation}, 2019, pp. 8691--8697.
  [Online]. Available: \url{http://arxiv.org/abs/1905.07447}
\BIBentrySTDinterwordspacing

\bibitem{Levine2018}
\BIBentryALTinterwordspacing
S.~Levine, P.~Pastor, A.~Krizhevsky, J.~Ibarz, and D.~Quillen, ``{Learning
  hand-eye coordination for robotic grasping with deep learning and large-scale
  data collection},'' \emph{International Journal of Robotics Research},
  vol.~37, no. 4-5, pp. 421--436, 2018. [Online]. Available:
  \url{http://arxiv.org/abs/1603.02199}
\BIBentrySTDinterwordspacing

\bibitem{Kleeberger2020a}
\BIBentryALTinterwordspacing
K.~Kleeberger, M.~V{\"{o}}lk, M.~Moosmann, E.~Thiessenhusen, F.~Roth,
  R.~Bormann, and M.~F. Huber, ``{Transferring Experience from Simulation to
  the Real World for Precise Pick-And-Place Tasks in Highly Cluttered
  Scenes},'' in \emph{International Conference on Intelligent Robots and
  Systems}, 2020. [Online]. Available:
  \url{http://www.bin-picking.ai/en/competition.html}
\BIBentrySTDinterwordspacing

\end{thebibliography}

\end{document}